\documentclass{article} % For LaTeX2e
\usepackage{iclr2025_conference,times}

\usepackage{amsmath,amsfonts,bm}

\def\eqref#1{equation~\ref{#1}}
\def\1{\bm{1}}

\DeclareMathAlphabet{\mathsfit}{\encodingdefault}{\sfdefault}{m}{sl}
\SetMathAlphabet{\mathsfit}{bold}{\encodingdefault}{\sfdefault}{bx}{n}

\usepackage{hyperref}
\usepackage{url}

\usepackage{booktabs}
\usepackage{multirow}
\usepackage{rotating}
\usepackage{colortbl}
\usepackage{xcolor}
\definecolor{lightgray}{gray}{0.9}
\usepackage{wrapfig}
\usepackage{xcolor}
\usepackage{hyperref}

\newcommand{\extractorlong}{\textsc{Vision Textualizer}}
\newcommand{\reasonerlong}{\textsc{Textual Reasoner}}
\newcommand{\extractor}{\textsc{Textualizer}}
\newcommand{\reasoner}{\textsc{Reasoner}}
\newcommand{\extractorshort}{\textsc{T}}
\newcommand{\reasonershort}{\textsc{R}}
\newcommand{\modelname}{\textsc{TextFlow}}

\newcommand{\textreptypeGraphviz}{\textsc{Graphviz}}
\newcommand{\textreptypeMermaid}{\textsc{Mermaid}}
\newcommand{\textreptypePlant}{\textsc{PlantUML}}

\title{Beyond End-to-End VLMs: \\Leveraging Intermediate Text Representations for Superior Flowchart Understanding}

\author{Junyi Ye$^\dagger$, Ankan Dash$^\dagger$, Wenpeng Yin$^\P$, Guiling Wang$^\dagger$\\
$^\dagger$New Jersey Institute of Technology;
$^\P$Pennsylvania State University\\
\texttt{\{jy394,ad892,guiling.wang\}@njit.edu}; \texttt{wenpeng@psu.edu}
}

\iclrfinalcopy % Uncomment for camera-ready version, but NOT for submission.
\begin{document}

\maketitle

\begin{abstract}
Flowcharts are typically presented as images, driving the trend of using vision-language models (VLMs) for end-to-end flowchart understanding. However, two key challenges arise: (i) \textit{Limited controllability}—users have minimal influence over the downstream task, as they can only modify input images, while the training of VLMs is often out of reach for most researchers. (ii) \textit{Lack of explainability}—it is difficult to trace VLM errors to specific causes, such as failures in visual encoding or reasoning. We propose \modelname, addressing aforementioned issues with two stages: (i) \extractorlong—which generates textual representations from flowchart images; and (ii) \reasonerlong—which performs question-answering based on the text representations. \modelname~offers three key advantages: (i) users can select the type of text representations (e.g., \textreptypeGraphviz, \textreptypeMermaid, \textreptypePlant), or further convert them into executable graph object to call tools, enhancing performance and \emph{controllability}; (ii) it \emph{improves explainability} by helping to attribute errors more clearly to visual or textual processing components; and (iii) it \emph{promotes the modularization} of the solution, such as allowing advanced LLMs to be used in the \reasoner~stage when VLMs underperform in end-to-end fashion. Experiments on the FlowVQA and FlowLearn benchmarks demonstrate \modelname's state-of-the-art performance as well as its robustness. All code is publicly available\footnote{GitHub: \url{https://github.com/JunyiYe/TextFlow}}.
\end{abstract}

\section{Introduction}

Flowcharts are extensively used to represent processes, algorithms, and workflows across a range of domains, including software engineering, business process modeling, and education. Accurate interpretation of flowcharts is essential for tasks such as automation, decision-making, and analysis. Given that flowcharts predominantly exist as images, the rise of large language models (LLMs) and large vision-language models (VLMs) has led to the use of VLMs for flowchart understanding in an end-to-end manner \citep{tannert-etal-2023-flowchartqa,pan2024flowlearnevaluatinglargevisionlanguage,singh-etal-2024-flowvqa}.

While end-to-end VLMs offer a straightforward approach to flowchart understanding, they exhibit two key limitations: (i) \textit{Limited controllability}—users have minimal capacity to improve performance, as they can only manipulate input images, while training VLMs is resource-intensive and often inaccessible to most researchers. (ii) \textit{Lack of explainability}—it is difficult to trace errors in VLM outputs to specific failures, whether in visual encoding, reasoning, or other stages.

To overcome these challenges, we introduce \modelname, a framework that decomposes flowchart understanding into two stages: (i) \extractorlong, which generates intermediate textual representations from flowchart images; and (ii) \reasonerlong, which performs question-answering (QA) based on these text representations. This dual-stage framework provides three distinct advantages: (i) \textbf{Controllability}—users can flexibly choose the type of text representation (e.g., \textreptypeGraphviz, \textreptypeMermaid, or \textreptypePlant) and convert them into executable graph objects for enhanced performance; (ii) \textbf{Explainability}—it improves error attribution by clarifying whether failures arise from visual or textual components; and (iii) \textbf{Modularity}—the framework allows for the use of more advanced LLMs in the \reasoner~stage if VLMs are inadequate for end-to-end flowchart understanding, restricting the VLM to the \extractor~stage only.

We evaluate \modelname~on the FlowVQA \citep{singh-etal-2024-flowvqa} and FlowLearn \citep{pan2024flowlearnevaluatinglargevisionlanguage} datasets, combining various open-source and closed-source VLMs/LLMs for the \extractorlong~and \reasonerlong~stages, respectively. Our experiments demonstrate that using Claude-3.5 \citep{anthropic2024} as the \extractor, alongside either Claude-3.5 or GPT-4o \citep{openai2024} as the \reasoner, significantly outperforms end-to-end flowchart understanding using Claude-3.5 or GPT-4o alone. Further analysis reveals that (i) \textreptypeGraphviz~is generally the most effective textual representation for flowcharts; (ii) \modelname~exhibits robustness across different task categories, flowchart sources, orientations, and sizes; and (iii) the \extractorlong~stage is more often the source of errors than the \reasoner~stage, particularly when using Claude-3.5 for both.

Our contributions are threefold:
% \begin{itemize}
%     \item Our \modelname~presents a novel and effective approach to leverage VLMs/LLMs in flowchart understanding: instead of treating the task end-2-end, breaking it into subtasks appears to be easier. It not only achieves new state-of-the-art (82.74 vs. 76.61) but also potentially provides inspiration for other visual tasks.
%     \item     We take the lead in introducing three textual formats—\textreptypeGraphviz, \textreptypeMermaid, and \textreptypePlant—for transforming flowcharts into structured text representations, facilitating the reasoning process.
%     \item Our detailed analysis provides key insights into solving the flowchart understanding problem, including identifying \textreptypeGraphviz~as the most effective representation, conducting a fine-grained analysis across different flowchart sources and task categories, and quantitatively comparing the roles of \extractorlong~and \reasonerlong~when using the same VLM.
% \end{itemize}

\textbullet\enspace Our \modelname~presents a novel and effective approach to leverage VLMs/LLMs in flowchart understanding: instead of treating the task end-2-end, breaking it into subtasks appears to be easier. It not only achieves new state-of-the-art (82.74 vs. 76.61) but also potentially provides inspiration for other visual tasks.

\textbullet\enspace We take the lead in introducing three textual formats—\textreptypeGraphviz, \textreptypeMermaid, and \textreptypePlant—for transforming flowcharts into structured text representations, facilitating the reasoning process.

\textbullet\enspace Our detailed analysis provides key insights into solving the flowchart understanding problem, including identifying \textreptypeGraphviz~as the most effective representation, conducting a fine-grained analysis across different flowchart sources and task categories, and quantitatively comparing the roles of \extractorlong~and \reasonerlong~when using the same VLM.

\section{Related Work}

\paragraph{End-2-End VQA for Flowchart/Diagrams}
For flowchart understanding, most prior work has focused on benchmark construction with different emphases. For example, FlowchartQA \citep{tannert-etal-2023-flowchartqa} emphasizes reasoning over geometric and topological features, FlowLearn \citep{pan2024flowlearnevaluatinglargevisionlanguage} focuses on synthetic and scientific flowcharts, IconQA \citep{lu2021iconqa} targets abstract diagrams that are rich in semantics rather than natural images, and FlowVQA \citep{singh-etal-2024-flowvqa} focuses on evaluating spatial reasoning, decision-making, and logical progression tasks.

Most of these works adopt end-to-end systems. For instance, \cite{lu2021iconqa} proposed Patch-TRM, which employs a pyramid cross-modal Transformer with diagram embeddings pre-trained on the Icon dataset, while \cite{singh-etal-2024-flowvqa} and \cite{pan2024flowlearnevaluatinglargevisionlanguage} explored various VLMs for their respective tasks.

\paragraph{Decomposing VQA into Multiple Steps}

Previous work in this area falls into two main categories: (i) \textbf{Image preprocessing}, where the decomposition involves processing flowchart images. For example, \cite{pan2024flowlearnevaluatinglargevisionlanguage} first identified visual components and applies OCR to the images. (ii) \textbf{Question decomposition}, where the focus is on breaking down questions into sub-reasoning steps. For instance, \cite{cao-jiang-2023-modularized} decomposed questions into sub-tasks and assigned them to suitable pre-trained models without adaptation. \cite{10.5555/3666122.3668594} compared human-written and model-generated question decompositions, while \cite{barezi2024disentanglingknowledgebasedvisualreasoning} decomposed multi-hop questions by determining the modality required, using a captioner for visual sub-questions and LLMs for textual ones. IdealGPT \citep{you2023idealgpt} employed LLMs to generate sub-questions, VLMs to provide sub-answers, and another LLM to reason and produce the final answer.

Our approach differs from these by focusing not on preprocessing flowchart images or decomposing questions, but on \textbf{decomposing the process} into two distinct stages. The first stage generates a novel textual representation, and the second transforms the original visual QA task into a textual QA task.

\section{Methodology}

\begin{figure*}[ht]
    \centering
    \includegraphics[width=\textwidth]{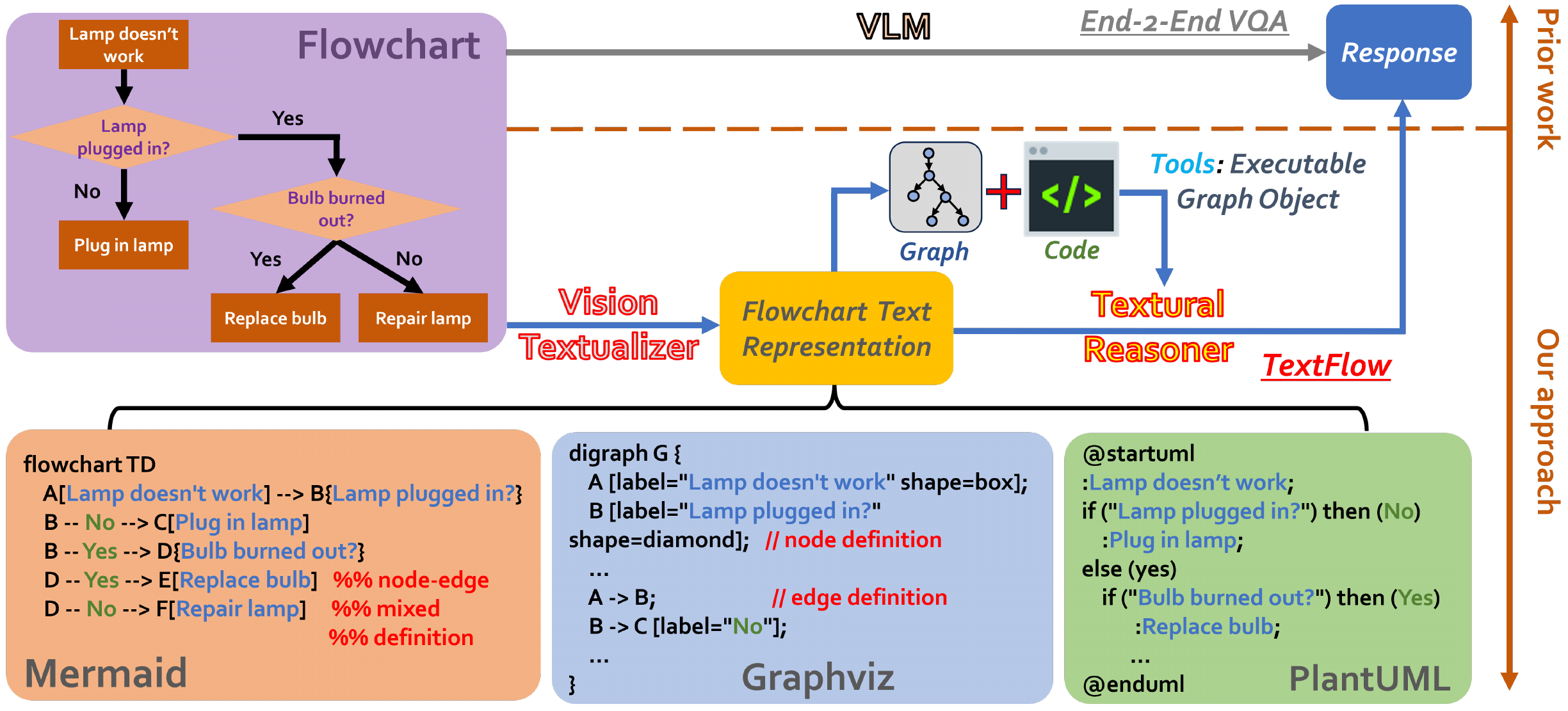}
    \caption{Our dual-stage \modelname~vs. prior work.}
    \label{fig:textflow}
\end{figure*}

As illustrated in Figure \ref{fig:textflow}, the approach \modelname~consists of two main stages: (1) \extractorlong: A VLM converts the visual components of the flowchart into a high-quality text representation. (2) \reasonerlong: An LLM or VLM based \reasonerlong~ can use tools and interpret the text representations to answer questions related to the flowchart’s logic and structure. By decoupling these tasks, \modelname~enables specialized models to focus on their respective strengths: \extractorlong~ process the visual elements, while \reasonerlong~ handle reasoning tasks, thus improving accuracy and flexibility. Next, we  elaborate on \extractorlong~(Section \ref{sec:extractor}) and \reasonerlong~(Section \ref{sec:reasoner}), respectively. Detailed prompts for both stages are provided in the Appendix.

\subsection{\extractorlong}\label{sec:extractor} 

\extractor~aims at extracting a structured textual representation from the flowchart image. This representation is critical because the accuracy and completeness of the extracted text determine the effectiveness of the reasoning stage.
We explore three  flowchart textualization formats (examples can be found in Figure \ref{fig:textflow}):

\paragraph{\textreptypeMermaid\protect\footnote{Mermaid: \url{https://mermaid.js.org/}}.} It uses a simple, link-based syntax, where nodes are connected via arrows. It is designed for ease of use, making it straightforward to define basic relationships and linear flows, suitable for simpler diagrams. Shapes are defined with different types of parentheses. Additionally, \textreptypeMermaid~often uses alphanumeric identifiers to label nodes, which simplifies counting nodes and edges and tracking direct connections.

\paragraph{\textreptypeGraphviz\protect\footnote{Graphviz: \url{https://graphviz.org/}}.} It defines nodes and edges separately, with nodes assigned attributes like labels and shapes, and edges specified using directional arrows. Unlike \textreptypeMermaid, which combines nodes and links in a single step, \textreptypeGraphviz’s approach is more structured and easier to follow. Due to their similar nature, converting between the two formats is straightforward. However, for more complex structures like loops or nested conditions, both \textreptypeMermaid~and \textreptypeGraphviz~representations can become less intuitive, as their simple syntax struggles to fully capture the intricacies of multi-step processes.

\paragraph{\textreptypePlant\protect\footnote{PlantUML: \url{https://plantuml.com/}}.} It takes a pseudocode-like approach, mimicking the structure of programming logic. Unlike \textreptypeMermaid~and \textreptypeGraphviz, it supports complex flow structures such as conditions, loops, and nesting, making it suitable for detailed and intricate flowcharts. However, its more elaborate syntax requires a deep understanding of process logic, and writing or maintaining flowcharts with complex topologies can be challenging. For instance, handling multiple nested loops with conditions may require significant restructuring of the flowchart logic, which complicates both the code writing and the accurate depiction of the topological relationships between components.

\subsection{\reasonerlong}\label{sec:reasoner}

At this stage, the text representation produced by \extractor~is forwarded to the \reasoner, which may consist of an LLM or VLM, to perform reasoning tasks guided by the flowchart's underlying logic. This stage highlights the \modelname~pipeline’s primary advantage over traditional VQA systems: \textit{controllability}—allowing the use of the plain text representation as generated by \extractor~or the further enhancement of its expressivity.

\textbullet\enspace\textbf{Plain Text Representation:} The generated text representation (in formats such as \textreptypeGraphviz, \textreptypeMermaid, or \textreptypePlant) is presented to the \reasoner~along with a question regarding the flowchart. The LLM/VLM processes the structured text to provide answers related to the flowchart's logic, decision paths, or conditional structures.

\textbullet\enspace\textbf{Executable Graph Object:} For improved reasoning, the user may opt to transform the text representation into an executable graph object through tool integration. This step is particularly effective for tasks requiring an understanding of topology, such as navigating loops, branches, and nested conditions, as it provides a more concrete structure for the LLM/VLM to analyze. This transformation enhances the reasoner’s accuracy in handling complex logical flows.

% \textbullet\textbf{Answer Generation:} Based on the graph structure or the plain text representations, the LLM/VLM generates an answer to the posed question. The use of an executable graph object allows the reasoner to handle topological reasoning more effectively, improving performance on questions involving complex, multi-step processes.

\section{Experiments}

\paragraph{Dataset.}
We utilize the \textbf{FlowVQA} dataset \citep{singh-etal-2024-flowvqa} due to its extensive diversity in i) \textit{flowchart sources}, including: \textsc{Code}, \textsc{Wiki} (step-by-step guides for daily tasks), and \textsc{Instruct} (instructions for DIY projects), and ii) \textit{task categories}:
% \begin{itemize}
%     \item Fact Retrieval ($T_1$): Focuses on extracting specific facts from flowchart nodes.
%     \item Applied Scenarios ($T_2$): Assesses how models apply flowchart logic in practical, real-world situations.
%     \item Flow Referential ($T_3$): Requires reasoning over flowchart subgraphs to perform forward or backward logic.
%     \item Topological ($T_4$): Evaluates understanding of structural metrics such as node count, edge count, shortest path, maximum in/out degree, and predecessor/successor relationships.
% \end{itemize}

\textbullet\enspace Fact Retrieval ($T_1$): Focuses on extracting specific facts from flowchart nodes.

\textbullet\enspace Applied Scenarios ($T_2$): Assesses how models apply flowchart logic in practical, real-world situations.

\textbullet\enspace Flow Referential ($T_3$): Requires reasoning over flowchart subgraphs to perform forward or backward logic.

\textbullet\enspace Topological ($T_4$): Evaluates understanding of structural metrics such as node count, edge count, shortest path, maximum in/out degree, and predecessor/successor relationships.

FlowVQA emphasizes spatial reasoning and understanding of decision-making flows within flowcharts, offering a robust resource to evaluate multimodal models' ability to interpret and process structured, graph-based visual information—a demanding task for many VLMs. For our experiment, we randomly sampled 200 flowcharts with a total of 2,005 QA pairs from the test set.

We also use the \textbf{FlowLearn} \citep{pan2024flowlearnevaluatinglargevisionlanguage} dataset to benchmark different systems on flowchart comprehension, focusing on the Simulated Flowcharts Subset, which features flowcharts generated with random words and links. Key tasks include OCR, True/False statements, description generation, and counting nodes/arrows, with emphasis on topological analysis as node labels lack semantic meaning. We evaluated 7 tasks from the test set, each with 100 flowchart/QA pairs. Besides the main results in Table \ref{tab:mainresults}, further analysis primarily focuses on FlowVQA, as it better reflects real-world scenarios.

% \begin{table}[t]
% \small
% \centering
% % \renewcommand{\arraystretch}{1.2} % Increase row spacing for better readability
% \begin{tabular}{c|lc}
% \toprule
% \multicolumn{2}{c}{\textbf{System}} & Overall\\
% \midrule
% \multirow{7}{*}{\parbox{1.8cm}{VQA\cite{singh-etal-2024-flowvqa}\\(end-to-end)}}
%  & Llama3.2-11B & 8.33 \\
%  & Llava-v1.6-110B & 42.69 \\
%  & Llama3.2-90B & 43.09 \\
%  & Qwen2-VL-7B & 53.42 \\
%  & Qwen2-VL-72B & 64.14 \\
%  & GPT-4o & 65.69 \\
%  & Claude3.5-Sonnet & 76.61 \\
%  % & paper 3 &    \\
% \midrule
% Ours& \extractor=\reasoner & \textbf{82.74} \\
% (dual-stage)& \extractor~$\neq$ \reasoner~& 82.19 \\

% \bottomrule
% \end{tabular}
% \caption{Flowchart QA evaluation. In the \extractor=\reasoner\ configuration, both the \extractor~and \reasoner~are Claude-3.5-Sonnet. In the \extractor~$\neq$~\reasoner\ setup, the most effective configuration is \extractor=GPT-4o and \reasoner=Claude-3.5-Sonnet.}
% \label{tab:mainresults}
% \end{table}

\begin{table}[t]
\small
\centering
\setlength{\tabcolsep}{2pt} % Adjust column padding
\begin{tabular}{c|l|cc}
\toprule
\textbf{} & \textbf{System} & FlowVQA & FlowLearn\\
\midrule
\multirow{7}{*}{\shortstack{VQA\\Baseline\\(end-to-end)\\\citep{singh-etal-2024-flowvqa}\\\citep{,pan2024flowlearnevaluatinglargevisionlanguage}}}  & Llama3.2-11B & 8.33 & - \\
 & Llava-v1.6-110B & 42.69 & - \\
 & Llama3.2-90B & 43.09 & - \\
 & Qwen2-VL-7B & 53.42 & - \\
 & Qwen2-VL-72B & 64.14 & - \\
 & GPT-4o & 65.69 & 60.29 \\
 & Claude3.5-Sonnet & 76.61 & 77.00\\
\midrule
Ours& GPT-4o & 80.10 & 72.83 \\
(dual-stage)& Claude3.5-Sonnet & \textbf{82.74} & \textbf{80.57}\\

\bottomrule
\end{tabular}
\caption{Flowchart of QA evaluation on two benchmarking datasets. The text representations used in the \modelname~pipeline are  \textreptypeGraphviz.}
\label{tab:mainresults}
\end{table}

% \begin{table*}[ht]
% \small
% \centering
% \caption{Performance Comparison of LVLMs on Graphviz, Mermaid, and Plantuml Extraction.}
% \label{tab:extraction_evaluation}
% \begin{tabular}{l|cc|cc|cc|cc|cc|cc}
% \hline
% \textbf{Model} & \multicolumn{4}{c|}{\textbf{Graphviz}} & \multicolumn{4}{c|}{\textbf{Mermaid}} & \multicolumn{4}{c}{\textbf{Plantuml}} \\
% \hline
% & \multicolumn{2}{c|}{\textbf{Node}} & \multicolumn{2}{c|}{\textbf{Edge}} & \multicolumn{2}{c|}{\textbf{Node}} & \multicolumn{2}{c|}{\textbf{Edge}} & \multicolumn{2}{c|}{\textbf{Node}} & \multicolumn{2}{c}{\textbf{Edge}} \\
% & Acc & F1 & Acc & F1 & Acc & F1 & Acc & F1 & Acc & F1 & Acc & F1 \\
% \hline
% GPT-4o & \textbf{97.21} & \textbf{98.47} & \textbf{88.29} & \textbf{92.87} & \textbf{98.18} & \textbf{99.00} & \textbf{90.26} & \textbf{94.32} &  &  &  &  \\
% Claude-3.5-Sonnet & 87.03 & 91.20 & 54.01 & 62.10 & 94.30 & 96.74 & 83.58 & 89.49 &  &  &  &  \\
% Qwen2-VL-72B & 80.27 & 86.78 & 60.58 & 69.01 & 69.67 & 73.10 & 57.35 & 63.87 &  &  &  &  \\
% Llama3.2-90B & 62.60 & 71.16 & 34.20 & 45.58 & 69.74 & 73.19 & 45.39 & 54.66 &  &  &  &  \\
% Llava-v1.6-110B & 42.40 & 51.18 & 21.38 & 28.97 & 37.09 & 46.84 & 15.13 & 22.57 &  &  &  &  \\
% Qwen2-VL-7B & 36.15 & 39.63 & 18.31 & 23.58 & 28.46 & 37.20 & 8.35 & 11.91 &  &  &  &  \\
% Llama3.2-11B & 40.44 & 49.21 & 12.56 & 18.57 & 10.33 & 12.49 & 4.33 & 5.58 &  &  &  &  \\
% \hline
% \end{tabular}
% \end{table*}

\begin{table*}[ht]
\small
\centering
\renewcommand{\arraystretch}{1.1} % Increase row spacing for better readability
\setlength{\tabcolsep}{4pt}
\begin{tabular}{l|ccc|ccc|ccc}
\hline
\textbf{Model} & \multicolumn{3}{c|}{\textit{\textreptypeGraphviz}} & \multicolumn{3}{c|}{\textit{\textreptypeMermaid}} & \multicolumn{3}{c}{\textit{\textreptypePlant}} \\
\hline
& Node & Link & Rendering & Node & Link & Rendering & Node & Link & Rendering \\
& F1 & F1 & Success Rate & F1 & F1 & Success Rate & F1 & F1 & Success Rate \\
\hline
GPT-4o & \textbf{0.98} & \textbf{0.93} & \textbf{100} & \textbf{0.99} & \textbf{0.94} & \textbf{100} & \textbf{0.97} & \textbf{0.88} & \textbf{100} \\
Claude3.5-Sonnet & 0.95 & 0.87 & \textbf{100} & 0.97 & 0.89 & 98 & 0.94 & 0.83 & 82 \\
Qwen2-VL-72B & 0.92 & 0.78 & \textbf{100} & 0.73 & 0.64 & 74 & 0.74 & 0.56 & 74 \\
Llama3.2-90B & 0.71 & 0.46 & 62 & 0.73 & 0.55 & 84 & 0.70 & 0.51 & 74 \\
Llava-v1.6-110B & 0.51 & 0.29 & 86 & 0.47 & 0.23 & 82 & 0.41 & 0.23 & 68 \\
Qwen2-VL-7B & 0.47 & 0.28 & 68 & 0.37 & 0.12 & 12 & 0.27 & 0.15 & 18 \\
Llama3.2-11B & 0.49 & 0.19 & 34 & 0.12 & 0.06 & 8 & 0.17 & 0.09 & 10 \\
\hline
\end{tabular}
\caption{\extractorlong~evaluation.}
\label{tab:extraction_evaluation}
\end{table*}

\paragraph{VLM\&LLM Explored.} Note that \extractor~can only use VLMs, but both VLMs and LLMs are applicable to \reasoner~.

\textbullet\enspace\textbf{VLMs:} Closed-source models: Claude 3.5 Sonnet and GPT-4o; open-source models: Qwen2-VL (7B and 72B) \citep{wang2024qwen2}, Llama3.2 (11B and 90B) \citep{llama3_2}, and Llava-v1.6-110B \citep{liu2024llavanext}. 
% Our LVLMs include cutting-edge models such as \textbf{GPT-4o} and \textbf{Claude 3.5-Sonnet}, both of which have demonstrated strong performance in multimodal tasks. We also evaluate Qwen2-VL models \cite{wang2024qwen2}, specifically \textbf{Qwen2-VL-7B} and \textbf{Qwen2-VL-72B}, which have set new benchmarks in visual understanding across multiple scales. Additionally, the Llama 3.2 models \cite{llama3_2}, with two sizes (\textbf{Llama 3.2-11B} and \textbf{Llama 3.2-90B}), exhibit competitive performance in visual reasoning, closely rivaling top models like Claude 3 Haiku and GPT-4o-mini. Furthermore, we incorporate \textbf{LLAVA-v1.6-110B}, the largest model in the LLAVA series, known for its strong capabilities in visual-language tasks.

\textbullet\enspace\textbf{LLMs:} Closed-source models: Claude 3.5 Sonnet and GPT-4o; open-source models: Qwen2.5 (7B, 14B, 32B, and 72B) \citep{qwen2.5}, Llama3.1 (8B and 70B) \citep{dubey2024llama}, Mixtral 8x22B \citep{mixtral2024}, Phi3.5 (MoE, Mini) \citep{microsoft2024}.

We accessed GPT-4o and Claude3.5-Sonnet via API, while all other open-source VLMs/LLMs were used through Hugging Face with their official default settings. Depending on the GPU memory requirements of each model, we utilized 1 to 4 Nvidia A100 GPUs. To ensure experimental accuracy and reproducibility, we employed a greedy decoding strategy (with temperature = 0) and fixed the maximum token length to 4096. In addition, to avoid the impact of image resolution on VQA performance, we used the highest resolution mode supported by each VLM.

\subsection{Main Results}
As the main results, we evaluate our system for the whole process of flowchart understanding (i.e., \extractor+\reasoner) as well as the quality of the first stage (i.e., merely \extractor).

\paragraph{\modelname~ (\extractor+\reasoner) evaluation.} It checks whether \modelname~can get superior flowchart understanding performance.

\textbullet\enspace\textbf{Baseline:} \cite{singh-etal-2024-flowvqa} and \cite{pan2024flowlearnevaluatinglargevisionlanguage} presented the prior state-of-the-art results by deploying VLMs end-2-end. 

\textbullet\enspace\textbf{Metric:} We use accuracy as the primary evaluation metric, measuring the proportion of correct answers among all responses to assess the model’s performance on QA tasks. Following prior work that employs LLMs as evaluators \citep{singh-etal-2024-flowvqa}, we use LLMs to assess each response; specifically, we employ GPT-4o in our setup for enhanced accuracy. A response is considered correct if it matches the expected answer. Each answer is evaluated three times, with the final determination based on a majority vote. To minimize excessive randomness and ensure reliable majority voting, we set the model’s temperature to 0.2, keeping responses stable yet varied enough for accurate assessment. Detailed prompt for evaluation is provided in the Appendix.
% \textcolor{red}{@Junyi: reorganize this paragraph: first directly say you use accuracy, then explain how to decide one instance is correct.} \textcolor{blue} {Junyi: updated.}

Table \ref{tab:mainresults} presents a comparison between \modelname~and previous end-to-end flowchart understanding systems on two benchmarking datasets. 
% The results clearly demonstrate that \modelname~consistently outperforms prior approaches by significant margins, regardless of whether \extractor~uses the same VLM as \reasoner. 
The results clearly demonstrate that top performing models (i.e. Claude 3.5 and GPT-4o) within \modelname~consistently outperform their performance in prior approach by significant margins.
Notably, Claude-3.5-Sonnet emerges as the most effective VLM, excelling in both end-to-end deployment and our dual-stage framework.

\paragraph{\extractor~Evaluation.} The experiment is conducted on a randomly selected set of 50 flowcharts in the FlowVQA dataset.

\textbullet\enspace\textbf{Generation of gold text representation:} Different target text representations pose unique challenges in obtaining the ground truth. For \textreptypeMermaid, we directly used the representations from the dataset. For \textreptypeGraphviz, which has a similar syntax, we generated the ground truth by parsing it with a script. However, for \textreptypePlant, due to the absence of conversion tools, we manually created the ground truth representations. Further details on the manual PlantUML ground truth creation can be found in the Appendix.

\textbullet\enspace\textbf{Metric:} We evaluate node and link extraction using the F1 score and assess the rendering success rate (whether the flowchart renders without syntax errors). For \textreptypeMermaid~and \textreptypeGraphviz, we convert text representations into Python graph objects using a parser, then add nodes and edges, and compute the F1 score via a script. For \textreptypePlant, due to its complex pseudocode-like syntax and lack of automated conversion tools, we perform manual annotation for accuracy assessment. Further details on the manual annotation process can be found in the Appendix.

The performance in Table \ref{tab:extraction_evaluation} illustrates that GPT-4o leads across all representations, with Claude 3.5 close behind, while Qwen2-VL-72B performs best among open-source models. Node extraction generally outperforms link extraction, with comparable results for Graphviz and Mermaid, but lower for PlantUML. Challenges include weaker edge extraction, where correctly identifying and connecting nodes with high in/out-degrees often leads to missing or incorrect links. Open-source models struggle with syntax errors, such as misuse of special characters and improper loop handling, where loops are mistakenly repeated indefinitely, highlighting their limitations in flowchart extraction tasks.
% \textcolor{red}{@Junyi: i) you mentioned manual annotation multiple times here, you should explain the manual process in detail in the appendix. ii) add a paragraph to discuss the Table \ref{tab:extraction_evaluation}}

\subsection{Analysis}
In addition to presenting the main results, we further answer the following four research questions.

\begin{table*}[ht!]
\small
\centering
\renewcommand{\arraystretch}{1.2} % Increase row spacing for better readability
\begin{tabular}{c|l|c|ccc|cccc}
\toprule
\multirow{2}{*}{\textbf{Input}} & \multirow{2}{*}{\textbf{Model}} & \multirow{2}{*}{\textbf{Total}} & \multicolumn{3}{c|}{\textbf{Data Source}} & \multicolumn{4}{c}{\textbf{Tasks}} \\
\cmidrule(lr){4-10} & &  & \textbf{Code} & \textbf{Instruct} & \textbf{Wiki} & \textbf{$T_{1}$} & \textbf{$T_{2}$} & \textbf{$T_{3}$} & \textbf{$T_{4}$} \\
\midrule
\textbf{Image} & GPT-4o (VQA) & 65.69 & 78.23 & 69.37 & 59.96 & 73.83 & 68.16 & 69.80 & 58.91 \\

\midrule
\multirow{11}{*}{\begin{sideways}\textbf{Graphviz}\end{sideways}} 
& Claude3.5-Sonnet & \textbf{\underline{82.19}} & \textbf{91.48} & \textbf{84.27} & \textbf{78.32} & \textbf{90.12} & \textbf{81.84} & \textbf{76.07} & \textbf{81.11} \\
& GPT-4o & \underline{80.10} & 87.07 & 83.94 & 75.92 & 88.15 & 81.34 & 72.65 & 78.75 \\
& Qwen2.5-32B & \underline{77.46} & 88.64 & 79.80 & 72.88 & 86.42 & 75.12 & 70.09 & 77.33 \\
& Llama3.1-70B & \underline{74.71} & 83.91 & 77.15 & 70.66 & 87.41 & 78.36 & 71.79 & 68.12 \\
& Qwen2.5-14B & \underline{74.51} & 87.38 & 77.48 & 69.10 & 81.98 & 72.39 & 68.09 & 74.62 \\
& Qwen2.5-72B & \underline{72.72} & 84.54 & 77.15 & 66.79 & 83.46 & 70.90 & 70.09 & 69.54 \\
& Mixtral-8x22B & \underline{67.03} & 76.03 & 69.54 & 63.01 & 86.17 & 73.13 & 66.38 & 55.25 \\
& Llama3.1-8B & 63.79 & 71.61 & 66.72 & 59.87 & 82.22 & 59.20 & 48.72 & 63.40 \\
& Qwen2.5-7B & 61.25 & 79.18 & 61.26 & 56.00 & 78.77 & 64.68 & 54.13 & 54.19 \\
& Phi3.5-MoE & 52.77 & 67.19 & 53.97 & 47.88 & 67.16 & 58.96 & 50.14 & 44.04 \\
& Phi3.5-Mini & 48.68 & 59.62 & 52.15 & 43.54 & 63.95 & 45.02 & 37.89 & 47.58 \\

\midrule
\multirow{11}{*} {\begin{sideways}\textbf{Mermaid}\end{sideways}} 
& Claude3.5-Sonnet & \underline{80.00} & 90.85 & 81.62 & 75.92 & 89.63 & 80.85 & 74.93 & 77.10 \\ 
& GPT-4o & \underline{77.81} & 88.33 & 81.46 & 72.69 & 88.89 & 79.10 & 74.93 & 73.08 \\
& Qwen2.5-32B & \underline{74.76} & 86.75 & 76.16 & 70.48 & 84.94 & 72.14 & 70.94 & 72.73 \\
& Llama3.1-70B & \underline{74.41} & 81.07 & 76.99 & 71.03 & 88.40 & 81.09 & 69.52 & 66.59 \\
& Qwen2.5-14B & \underline{73.37} & 85.17 & 74.01 & 69.56 & 85.93 & 72.89 & 72.08 & 68.12 \\
& Qwen2.5-72B & \underline{70.67} & 86.12 & 72.52 & 65.13 & 84.69 & 67.91 & 70.09 & 65.53 \\
& Llama3.1-8B & \underline{66.03} & 74.76 & 66.39 & 63.28 & 86.67 & 67.91 & 61.82 & 57.02 \\
& Mixtral-8x22B & 63.89 & 74.45 & 64.57 & 60.42 & 85.43 & 76.37 & 67.24 & 46.28 \\
& Qwen2.5-7B & 63.44 & 74.45 & 63.58 & 60.15 & 84.20 & 63.93 & 57.83 & 55.61 \\
& Phi3.5-MoE & 54.71 & 67.19 & 52.98 & 52.03 & 73.83 & 64.18 & 54.42 & 41.20 \\
& Phi3.5-Mini & 54.51 & 62.46 & 54.47 & 52.21 & 72.84 & 59.70 & 51.85 & 44.39 \\

\midrule
\multirow{11}{*}{\begin{sideways}\textbf{PlantUML}\end{sideways}} 
& Claude3.5-Sonnet & \underline{70.17} & 78.55 & 73.01 & 66.14 & 84.69 & 79.10 & 68.95 & 59.50 \\
& GPT-4o & \underline{66.83} & 76.03 & 70.03 & 62.36 & 85.93 & 73.63 & 68.95 & 53.60 \\
& Qwen2.5-32B & 64.59 & 74.13 & 67.05 & 60.42 & 82.47 & 73.13 & 68.95 & 50.18 \\
& Llama3.1-70B & 61.85 & 67.19 & 63.41 & 59.41 & 84.20 & 78.36 & 69.23 & 40.26 \\
& Qwen2.5-72B & 61.00 & 68.14 & 62.42 & 58.12 & 80.25 & 66.17 & 68.95 & 46.04 \\
& Qwen2.5-14B & 60.70 & 67.51 & 61.59 & 58.21 & 82.96 & 69.65 & 66.38 & 43.45 \\
& Mixtral-8x22B & 57.41 & 62.15 & 57.78 & 55.81 & 82.96 & 75.87 & 63.82 & 33.77 \\
& Llama3.1-8B & 53.57 & 58.99 & 55.63 & 50.83 & 83.95 & 69.65 & 56.13 & 30.34 \\
& Qwen2.5-7B & 52.92 & 60.88 & 53.81 & 50.09 & 82.22 & 61.44 & 56.70 & 33.29 \\
& Phi3.5-MoE & 51.47 & 57.41 & 50.33 & 50.37 & 76.79 & 66.42 & 55.27 & 30.70 \\
& Phi3.5-Mini & 45.14 & 52.05 & 45.03 & 43.17 & 71.85 & 57.71 & 45.87 & 26.09 \\
\bottomrule
\end{tabular}
\caption{Effectiveness of text representations. \extractor: GPT-4o; \reasoner: various LLMs. The underlined results indicate that the experiment outperforms the VQA baseline.}
\label{tab:textrepresentation}
\end{table*}

\paragraph{$\mathcal{Q}_1$: Which text representation is the most effective?}  
To address this question, we fix the \extractor~as GPT-4o to generate different text representations and then evaluate the performance of various VLMs/LLMs as \textsc{Reasoners} on those representations.

As shown in Table \ref{tab:textrepresentation}, \textreptypeGraphviz~proves to be the most effective text representation for flowchart understanding overall. While \textreptypePlant~performs the worst, it still surpasses the end-to-end VQA baseline in most cases. This underscores the effectiveness of our approach’s core concept: generating intermediate text representations.

\paragraph{$\mathcal{Q}_2$: How robust our dual-stage pipeline is?} We study robustness from three dimensions: i) varying VLM/LLM choices in the two stages; ii) flowchart orientation; iii) flowchart size (i.e., \#node).

\textbullet\enspace\textbf{VLM/LLM choices in \extractor~and \reasoner.}  
The upper half of Table \ref{tab:robustness1} compares the performance of different VLMs in the \modelname~pipeline and the VQA pipeline. In most cases, \modelname~outperforms VQA. Notably, GPT-4o and Claude 3.5 emerge as the top performers, showcasing strong vision-textualization and reasoning capabilities. Additionally, \modelname~demonstrates strong robustness, and when the reasoning ability of VLMs is insufficient, performance can be improved by utilizing a stronger \reasoner, such as GPT-4o.

In the lower half of Table \ref{tab:robustness1}, we observe this flexibility in action. When GPT-4o is fixed as the \extractor, a variety of LLMs can act as high-performing \textsc{Reasoners}. Even smaller models, such as Qwen2.5-14B and 32B, demonstrate strong performance in this setup. Moreover, when replacing the gold representation with text extracted by GPT-4o, most \textsc{Reasoners}, particularly in Graphviz and Mermaid formats, show significant improvement.

% Table \ref{tab:robustness1}. \textcolor{red}{@Junyi: discuss Table 4 here; focusing on how robust your system is}

\begin{table*}[ht]
\small
\centering
\renewcommand{\arraystretch}{1.2} % Increase row spacing for better readability
\setlength{\tabcolsep}{2pt} % Adjust column padding
\begin{tabular}{lc|cccccc}

\bottomrule
\toprule
\multicolumn{8}{c}{\extractor~Varies}\\
\toprule
\multirow{2}{*}{Model=VLM} & \multirow{2}{*}{VQA} & \multicolumn{3}{c|}{\extractorshort=\reasonershort=Model} &\multicolumn{3}{c}{\extractorshort=Model; \reasonershort=GPT-4o} \\\
& & \textbf{Graphviz} & \textbf{Mermaid} & \textbf{PlantUML} & \textbf{Graphviz} & \textbf{Mermaid} & \textbf{PlantUML} \\
\midrule
 Claude3.5-Sonnet & \textbf{76.61} & \textbf{\underline{82.74}} & \textbf{\underline{82.04}} & \textbf{71.22} & \underline{77.66} & \underline{\textbf{79.95}} & 66.43 \\
 GPT-4o & 65.69 & \underline{80.10} & \underline{77.81} & \underline{66.83} & \underline{\textbf{80.10}} & \underline{77.81} & \underline{\textbf{66.83}} \\
 Qwen2-VL-72B & 64.14 & \underline{64.64} & 62.19 & 57.91 & \underline{76.11} & \underline{75.26} & \underline{64.54} \\
 Qwen2-VL-7B & 53.42 & 48.43 & 49.93 & 46.18 & \underline{59.55} & \underline{60.35} & \underline{58.80} \\
 Llama3.2-90B & 43.09 & \underline{53.97} & \underline{62.29} & \underline{49.93} & \underline{56.96} & \underline{64.09} & \underline{52.22} \\
 Llava-v1.6-110B & 42.69 & \underline{43.09} & \underline{43.24} & 39.50 & \underline{49.08} & \underline{45.94} & \underline{44.09} \\
 Llama3.2-11B & 8.33 & \underline{39.80} & \underline{41.40} & \underline{34.21} & \underline{50.07} & \underline{50.47} & \underline{46.23} \\

\bottomrule

\toprule
\multicolumn{8}{c}{\reasoner~Varies}\\
\toprule
\multirow{2}{*}{Model=LLM} & \multirow{2}{*}{VQA} & \multicolumn{3}{c|}{\extractorshort=GPT-4o; \reasonershort=Model} &\multicolumn{3}{c}{\extractorshort=Gold; \reasonershort=Model} \\\
& & \textbf{Graphviz} & \textbf{Mermaid} & \textbf{PlantUML} & \textbf{Graphviz} & \textbf{Mermaid} & \textbf{PlantUML} \\
\midrule
Claude-3-5-Sonnet & 70.00 & \textbf{83.03} & \textbf{85.48} & \textbf{71.57} & \textbf{90.59} & \textbf{92.64} & \textbf{74.64} \\
GPT-4o & \textbf{72.00} & 80.98 & 82.21 & 71.37 & 87.12 & 86.09 & 70.35 \\
Qwen2.5-32B & - & 79.14 & 79.96 & 69.53 & 85.48 & 84.46 & 66.26 \\
Llama-3.1-70B & - & 79.55 & 76.89 & 64.01 & 85.89 & 80.98 & 66.26 \\
Qwen2.5-14B & - & 78.32 & 78.73 & 64.01 & 84.46 & 81.39 & 65.03 \\
Qwen2.5-72B & - & 74.03 & 74.23 & 65.24 & 78.73 & 78.94 & 64.62 \\
Mixtral-8x22B & - & 64.62 & 66.87 & 59.51 & 66.26 & 68.71 & 58.49 \\
Llama-3.1-8B & - & 70.96 & 65.64 & 56.24 & 73.21 & 71.17 & 53.78 \\
Qwen2.5-7B & - & 67.08 & 63.39 & 54.19 & 69.33 & 66.87 & 56.65 \\
Phi-3.5-MoE & - & 57.87 & 56.85 & 56.03 & 57.67 & 58.08 & 53.99 \\
Phi-3.5-mini & - & 56.03 & 50.92 & 46.42 & 57.87 & 51.74 & 44.79 \\

\bottomrule
\end{tabular}
\caption{Framework robustness by choosing different VLMs/LLMs for \extractor~(\extractorshort) and \reasoner~(\reasonershort). The underlined results indicate that the experiment outperforms the VQA baseline.}
\label{tab:robustness1}
\end{table*}

% \begin{table}[ht]
% \small
% \centering
% \renewcommand{\arraystretch}{1.2} % Increase row spacing for better readability
% \setlength{\tabcolsep}{5pt} % Adjust column padding
% \begin{tabular}{lccc}
% \toprule
% \textbf{Configuration} & \textbf{Top-Down} & \textbf{Bottom-UP} & \textbf{Difference} \\
% \midrule
% Graphviz & 80.10 & 74.86 & 5.24 \\
% Mermaid  & 77.81 & 72.82 & 4.99 \\
% PlantUML & 66.83 & 64.34 & 2.49 \\
% VQA      & 65.69 & 59.25 & 6.44 \\
% \bottomrule
% \end{tabular}
% \caption{Comparison of GPT-4o’s Performance on Top-Down (Normal) and Bottom-Up (Reversed) Flowchart Configurations, Showing Accuracy Differences. \textcolor{red}{@Junyi: use bar figure to denote this table. Removing the "Difference" col}}
% \label{tab:reverse_study}
% \end{table}

\begin{figure}[t]
    \centering
    \includegraphics[width=0.65\textwidth]{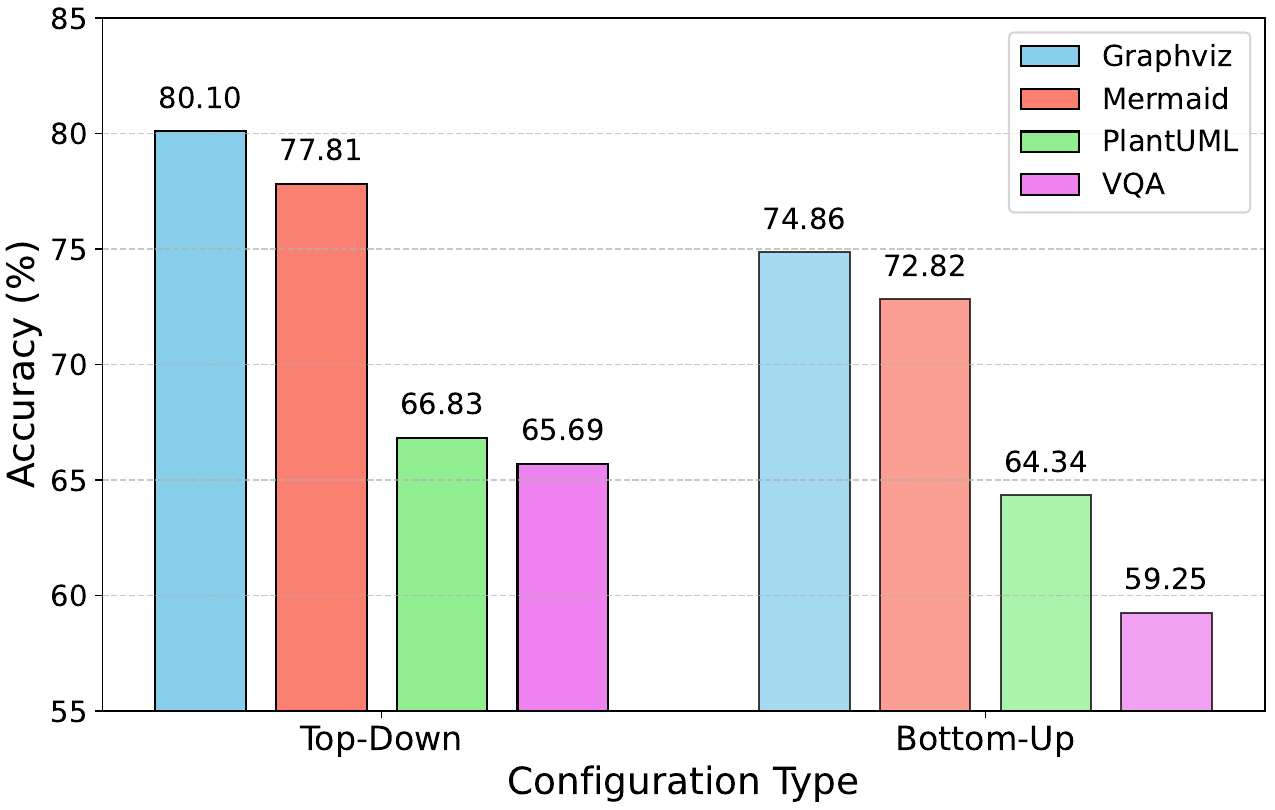}
    \caption{Comparison of GPT-4o’s performance on Top-Down and Bottom-Up flowchart configurations.}
    \label{fig:reverse_study}
\end{figure}

\textbullet\enspace\textbf{Impact of Flowchart Orientation.}
Figure \ref{fig:reverse_study} compares model accuracy between top-down (normal) and bottom-up (reversed) flowchart configurations. Across all representations (\textreptypeGraphviz, \textreptypeMermaid, and \textreptypePlant), the top-down configuration consistently yields higher accuracy, with differences ranging from 2.49\% for \textreptypePlant~ to 5.24\% for \textreptypeGraphviz. The VQA baseline shows the largest discrepancy, with a 6.44\% drop in accuracy in the bottom-up configuration. These results indicate that while reversing the flow direction affects performance across models, the \modelname~pipeline demonstrates better robustness than VQA in maintaining accuracy under reversed conditions. This robustness suggests that \modelname’s structured representation approach enables more flexible adaptation to variations in flowchart orientation.

\begin{figure}[t]
    \centering
    \includegraphics[width=0.65\textwidth]{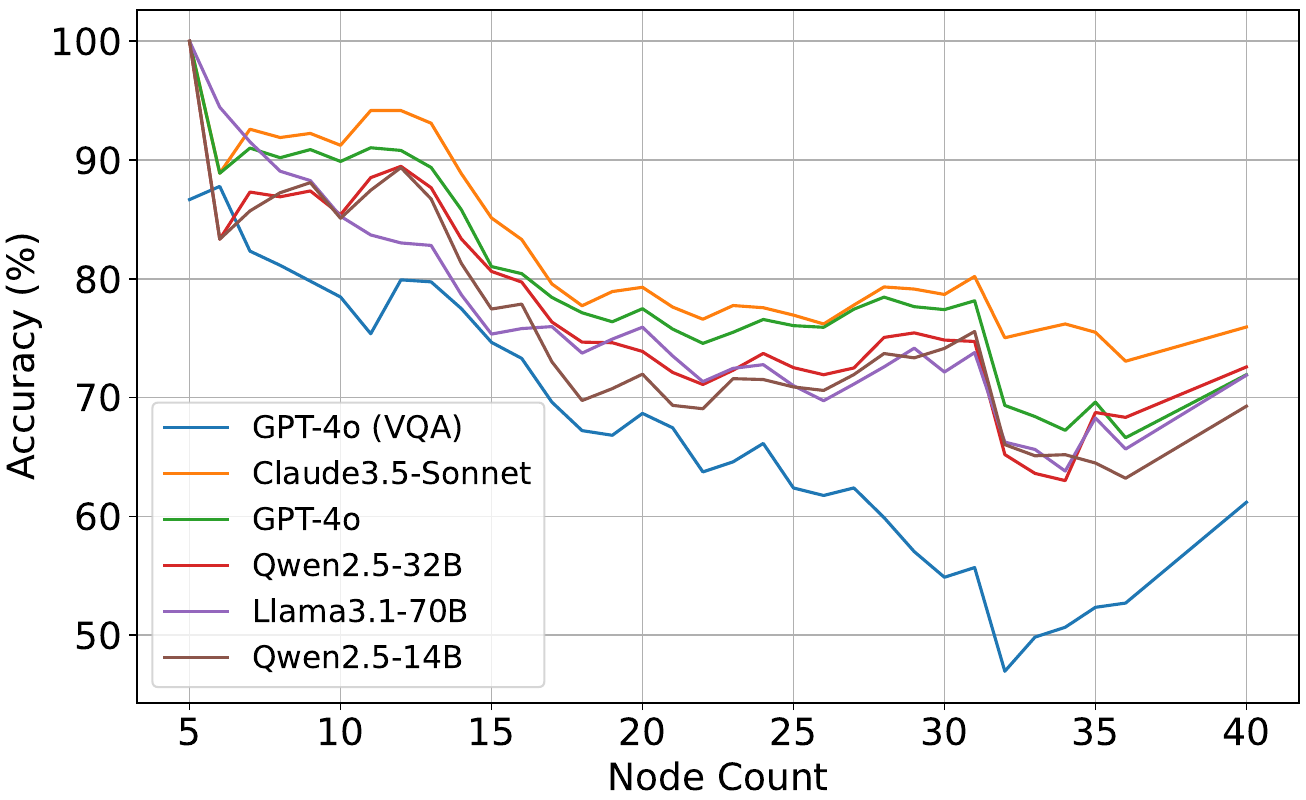}
    \caption{Accuracy comparison by node count across various models with a rolling average. VQA and the top 5 performing \textsc{\textsc{Reasoners}} on \modelname~using extracted Mermaid in Table \ref{tab:textrepresentation} are compared.}
    \label{fig:accuracy_node_count}
\end{figure}

\textbullet\enspace\textbf{Effect of Flowchart Size (e.g., \#Node).}
Figure \ref{fig:accuracy_node_count} explores how node count affects model accuracy for VQA and \modelname. Accuracy generally declines as node count increases, highlighting the challenge of processing more complex flowcharts. However, \modelname~demonstrates superior resilience compared to VQA, maintaining higher accuracy across increasing node counts. This suggests that the structured extraction and reasoning process in \modelname~enables models to handle complexity more effectively, whereas the VQA baseline struggles as flowcharts become denser and more intricate.

Together, these findings show that \modelname~enhances robustness to both orientation changes and increasing complexity, making it better suited for reliable flowchart reasoning across varied and challenging conditions.

\paragraph{$\mathcal{Q}_3$: Given that our dual-stage framework offers flexibility in controlling flowchart representations, how effective is it to enhance these representations using external tools?}

\begin{table}[t]
\small
\centering
\renewcommand{\arraystretch}{1.2} % Increase row spacing for better readability
\setlength{\tabcolsep}{6pt} % Adjust column padding
\begin{tabular}{lccccc}
\toprule
\textbf{Method} & \textbf{Overall} & \textbf{$T_1$} & \textbf{$T_2$} & \textbf{$T_3$} & \textbf{$T_4$} \\
\midrule
\multicolumn{6}{c}{\textreptypeGraphviz} \\
\midrule
Base & 80.10 & 88.15 & 81.34 & 72.65 & 78.75 \\
Tool & 80.79 & 84.94 & 80.60 & 70.37 & 83.23 \\
Base\textsuperscript{Gold}  & 85.39 & 90.12 & 79.35 & 76.35 & 89.73 \\
Tool\textsuperscript{Gold} & \textbf{88.23} & 86.91 & 79.35 & 72.08 & \textbf{99.76} \\
\midrule
\multicolumn{6}{c}{\textreptypeMermaid} \\
\midrule
Base & 77.81 & 88.89 & 79.10 & 74.93 & 73.08 \\
Tool & 78.89 & 86.91 & 79.35 & 69.52 & 78.87 \\
Base\textsuperscript{Gold}  & 85.49 & 90.37 & 78.11 & 78.92 & 89.37 \\
Tool\textsuperscript{Gold}  & \textbf{88.83} & 89.88 & 79.85 & 70.94 & \textbf{100.00} \\
\bottomrule
\end{tabular}
\caption{GPT-4o' accuracy on Graphviz and Mermaid representations, comparing base and tool-assisted settings on extracted and Gold representations.}
\label{tab:tool_evaluation}
\end{table}

We explore the impact of tool-assisted methods on enhancing text representations for improved model accuracy across various tasks, particularly topological ones. By converting flowchart representations from Mermaid and Graphviz code into Python-executable graph objects, we enrich the text representation with additional graph functions, such as node and edge counts, retrieving successors and predecessors, calculating shortest paths, and identifying nodes with maximum in-degree and out-degree. These tools provide structured information, enhancing the model's ability to reason more precisely in graph-based scenarios across the FlowVQA dataset. Details of the tools are provided in the Appendix.

As shown in Table \ref{tab:tool_evaluation}, tool-assisted methods significantly boost GPT-4o’s accuracy, particularly for topological tasks ($T_4$), where structured graph functions lead to near-perfect accuracy with Gold representation (99.76\% for Graphviz, 100\% for Mermaid). While these methods enhance overall performance for both Graphviz and Mermaid representations—achieving the highest accuracy (88.23\% for Graphviz, 88.83\% for Mermaid)—their impact is most pronounced in topological tasks. For other tasks ($T_1$-$T_3$), which may require multi-step reasoning or broader flowchart understanding, tool-assisted methods provide more limited benefits.

\paragraph{$\mathcal{Q}_4$: If \modelname~makes mistakes in the end, it is more likely due to the failure of \extractor~or \reasoner?} We explain the errors in three sources: i) \extractor~is correct, \reasoner~is incorrect; ii) \extractor~is incorrect, \reasoner~is correct (i.e., correctly reflect the facts in the text representations); iii) \extractor~is incorrect, \reasoner~is incorrect.

\begin{figure}[t]
    \centering
    \includegraphics[width=0.5\textwidth]{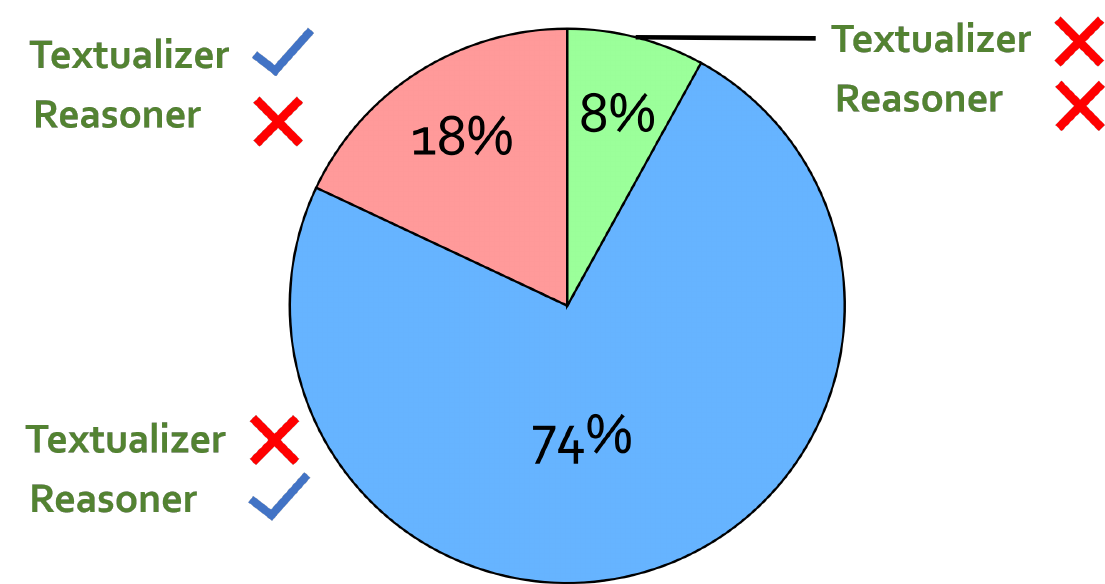}
    \caption{Error analysis for percentage of errors attributed to each category in Claude 3.5 Sonnet.}
    \label{fig:error_analysis}
\end{figure}

In our analysis of 50 randomly selected error cases from Claude-3.5, as shown in Figure \ref{fig:error_analysis}, the majority of errors were attributed to \extractor~issues. Specifically, errors frequently occurred in decision nodes where multiple links entered or exited, indicating that complex node structures pose challenges to accurate extraction. Another prevalent source of error was the misinterpretation of node labels due to unintended rewrites by the \extractor, leading to inaccuracies in extraction. Conversely, the \reasoner~component demonstrated a high degree of reliability, as evidenced by the lower error rate in reasoning tasks. These findings suggest that improvements in extraction quality, particularly in handling decision nodes and preserving label integrity, could unlock greater potential for our \modelname~model's overall performance.


\section{Conclusion}

This work introduces \modelname, a dual-stage framework that leverages VLMs/LLMs for flowchart understanding by breaking the process down into \extractorlong~and \reasonerlong. Our experiments on the two benchmarks demonstrate its state-of-the-art performance, and detailed analysis reveals that \textreptypeGraphviz~is the most effective text representation for flowcharts. Additionally, the system remains robust regardless of flowchart orientation and scale.
\clearpage
\section*{Limitations}

Despite the progress made with \modelname, several limitations remain, particularly in extraction accuracy, generalizability, and reasoning capabilities.

\begin{enumerate}
    \item \textbf{Extraction Accuracy and Representation Fidelity}: \modelname~depends on accurately extracting nodes and edges from flowcharts. In complex or noisy diagrams, errors in extraction can reduce reasoning accuracy. While current \extractorlong~perform well, they struggle with subtle layout and style variations, limiting \modelname's adaptability to diverse flowchart formats.
    
    \item \textbf{Lack of Diverse High-Quality Datasets}: \modelname's evaluation is limited by the scarcity of diverse, high-quality flowchart datasets. Existing datasets, like FlowVQA, mostly feature standard styles such as Mermaid. This restricts the assessment of TextFlow's performance on complex, real-world flowcharts. More varied datasets are needed to fully test its generalization abilities.
    
    \item \textbf{Limited Generalizability to Complex or Nested Diagrams}: \modelname~is designed for structured flowcharts but struggles with more complex diagrams, such as dependency graphs, Gantt charts, or those with nested structures or embedded images. These elements are challenging to extract and interpret accurately, requiring more advanced extraction methods to handle such complexity.
    
    \item \textbf{Dependence on Domain Knowledge and External Documents}: Some flowcharts require domain-specific knowledge or links to other documents for correct interpretation. In these cases, \modelname~may need to be integrated with retrieval-augmented generation (RAG) techniques or external knowledge bases to enhance its reasoning capabilities.
    
    \item \textbf{Reliance on External Graph Processing Tools}: \modelname~improves reasoning through the use of external graph-processing tools, but this increases system complexity and may lead to compatibility issues with future datasets. Reducing reliance on these tools and improving the system's self-sufficiency is a key area for future work.
\end{enumerate}

\clearpage

\bibliography{iclr2025_conference}
\bibliographystyle{iclr2025_conference}

\appendix
\clearpage
\onecolumn
\section{Prompt Details}
\label{sec
}

This section provides the details of the prompts used in the experiments.

\subsection{Prompt 1: Flowchart to Mermaid}

\textbf{Task:} Convert the provided flowchart into a Mermaid representation.\\

\noindent \textbf{Prompt template:}

\begin{verbatim}
Generate the Mermaid code for the provided flowchart.

Here is an example:
```mermaid
flowchart TD
    A(["Start"]) --> B[/"Receive 'arr' and 'n'"/]
    B --> C["Initialize loop index 'i' to 0"]
    C --> D{"Check if arr[i] == i"}
    D -->|"Yes"| E[/"Return index 'i' as fixed point"/]
    E --> F(["End"])
    D -->|"No"| G["Increment 'i'"]
    G --> H{"i < n"}
    H -->|"Yes"| D
    H -->|"No"| I[/"Return -1 as no fixed point found"/]
    I --> F
```
{image}

\end{verbatim}

\clearpage
\subsection{Prompt 2: Flowchart to Graphviz}

\textbf{Task:} Convert the provided flowchart into a Graphviz representation.\\

\noindent \textbf{Prompt template:}

\begin{verbatim}
Generate the Graphviz code for the provided flowchart.

Here is an example:
```dot
digraph G {
    A [label="Start" shape=ellipse];
    B [label="Receive 'arr' and 'n'" shape=parallelogram];
    C [label="Initialize loop index 'i' to 0" shape=box];
    D [label="Check if arr[i] == i" shape=diamond];
    E [label="Return index 'i' as fixed point" shape=parallelogram];
    F [label="End" shape=ellipse];
    G [label="Increment 'i'" shape=box];
    H [label="i < n" shape=diamond];
    I [label="Return -1 as no fixed point found" shape=parallelogram];

    A -> B;
    B -> C;
    C -> D;
    D -> E [label="Yes"];
    E -> F;
    D -> G [label="No"];
    G -> H;
    H -> D [label="Yes"];
    H -> I [label="No"];
    I -> F;
}
```
{image}

\end{verbatim}

\clearpage
\subsection{Prompt 3: Flowchart to PlantUML}

\textbf{Task:} Convert the provided flowchart into a PlantUML representation.\\

\noindent \textbf{Prompt template:}

\begin{verbatim}
Generate the PlantUML code for the provided flowchart.

Here is an example:
```plantuml
@startuml
start
:Receive 'arr' and 'n';
:Initialize loop index 'i' to 0;

while (i < n?) is (Yes)
    if (Check if arr[i] == i?) then (Yes)
        :Return index 'i' as fixed point;
        stop
    else (No)
        :Increment 'i';
    endif
endwhile (No)
:Return -1 as no fixed point found;
stop
@enduml
```
{image}

\end{verbatim}

\subsection{Prompt 4: Reasoning}

\textbf{Task:} Answer question based on the flowchart text representation.\\

\noindent \textbf{Prompt template:}

\begin{verbatim}
{Mermaid_code/Graphviz_code/PlantUMl_code}

Question: {question}
Answer: 
\end{verbatim}

\clearpage
\subsection{Prompt 5: Reasoning with Tools}

\textbf{Task:} Answer question based on the flowchart text representation with tools.\\

\noindent \textbf{Prompt template:}

\begin{verbatim}
{Mermaid_code/Graphviz_code/PlantUMl_code}

Question: {question}
Answer: 
{tools}
\end{verbatim}

\subsection{Prompt 6: Evaluation}

\textbf{Task:} Evaluate whether the model's answer is correct.\\

\noindent \textbf{Prompt template:}

\begin{verbatim}
Task: Verify if the provided answer is correct based on the given
ground truth.
    
You are given a question, an answer and the ground truth. Your
task is to determine whether the provided answer matches the 
ground truth. Output "Correct" if the answer matches ground truth, 
otherwise output "Incorrect".

Question: {question}

Answer: {answer}

Ground Truth: {ground_truth}
\end{verbatim}

\newpage
\section{Implemented Tools in the \modelname~Framework}

This section presents the tools implemented in our framework, including function names, their input and output parameters, and a brief definition.

\subsection{Functions Overview}

\begin{itemize}
    \item \textbf{get\_number\_of\_nodes()}
    \begin{description}
        \item[Input:] None
        \item[Output:] Returns the number of nodes in the flowchart.
    \end{description}

    \item \textbf{get\_number\_of\_edges()}
    \begin{description}
        \item[Input:] None
        \item[Output:] Returns the total number of edges in the flowchart.
    \end{description}

    \item \textbf{get\_direct\_successors(node\_description)}
    \begin{description}
        \item[Input:] \texttt{node\_description} - A string representing the description of a node.
        \item[Output:] A list of descriptions for the direct successors of the given node.
    \end{description}

    \item \textbf{get\_direct\_predecessors(node\_description)}
    \begin{description}
        \item[Input:] \texttt{node\_description} - A string representing the description of a node.
        \item[Output:] A list of descriptions for the direct predecessors of the given node.
    \end{description}

    \item \textbf{get\_shortest\_path\_length(start\_node\_description, end\_node\_description)}
    \begin{description}
        \item[Input:] \texttt{start\_node\_description}, \texttt{end\_node\_description} - Strings representing the descriptions of the start and end nodes.
        \item[Output:] An integer representing the number of edges in the shortest path between the two nodes. Returns -1 if no path is found.
    \end{description}

    \item \textbf{get\_max\_indegree()}
    \begin{description}
        \item[Input:] None
        \item[Output:] Returns the maximum indegree (number of incoming edges) for any node in the flowchart.
    \end{description}

    \item \textbf{get\_max\_outdegree()}
    \begin{description}
        \item[Input:] None
        \item[Output:] Returns the maximum outdegree (number of outgoing edges) for any node in the flowchart.
    \end{description}
\end{itemize}

\clearpage
\section{Manual Annotation Process}

This section explains how manual annotation is conducted using PlantUML to represent ground truth flowcharts. The process involves two annotators: one writes and the other verifies PlantUML code, ensuring the resulting flowchart accurately reflects the ground truth.

\subsection{Writing PlantUML Ground Truth Representation}

The manual annotation of a flowchart using PlantUML follows these steps:

\begin{enumerate}
    \item \textbf{First Annotator:} The first annotator manually writes the PlantUML code based on the ground truth flowchart. This code must precisely capture the structure and relationships between nodes and links as presented in the original flowchart.
    
    \begin{itemize}
        \item \textbf{Nodes:} The annotator writes PlantUML code that accurately reflects the text or description for each node in the flowchart. The text must be an exact match to what is shown in the ground truth flowchart.
        
        \item \textbf{Links:} The connections between nodes, including the direction of arrows and any attributes such as dashed or solid lines, should be represented in the PlantUML code to mirror the original flowchart.
    \end{itemize}
    
    \item \textbf{Second Annotator:} The second annotator verifies the accuracy of the PlantUML code by comparing the generated flowchart with the ground truth. The review process includes:
    
    \begin{itemize}
        \item \textbf{Node Verification:} Ensuring that all node text in the generated flowchart matches the text in the ground truth flowchart. The layout or shape of the nodes is not considered; only the textual content matters.
        
        \item \textbf{Link Verification:} Checking that the links between nodes, along with their direction and any attributes, match the original ground truth flowchart.
    \end{itemize}
    
    If any discrepancies are identified, both annotators collaborate to resolve them, ensuring that the final PlantUML code is a faithful representation of the ground truth.
\end{enumerate}

\subsection{Evaluating Extraction Quality of PlantUML}

Once the PlantUML code has been generated by the \extractor, the quality of the generated flowchart image is evaluated by comparing it to the ground truth. The evaluation focuses on two key metrics: the F1 score for node and link extraction, and the rendering success rate.

\begin{enumerate}
    \item \textbf{F1 Score for Node and Link Extraction:}
    
    \begin{itemize}
        \item \textbf{Node Extraction:} The generated nodes are compared to those in the ground truth flowchart. The comparison is based solely on the text of the nodes, without considering shape or layout. The F1 score is calculated based on how accurately the node descriptions are extracted.
        
        \item \textbf{Link Extraction:} The links between nodes are verified, ensuring that both the connections and arrow directions match the ground truth. Link attributes, such as arrow directions or labels, must also be correctly represented. The F1 score reflects the precision and accuracy of this link extraction.
    \end{itemize}
    
    \item \textbf{Rendering Success Rate:} The rendering success rate checks whether the PlantUML-generated image contains any syntax errors. The annotators manually verify that the PlantUML code is free of syntax errors. If any syntax issues arise, the flowchart fails to render.
 
\end{enumerate}

\subsection{Consistency with Script-Based Tools}

The manual annotation process is designed to meet the same standards as automated tools for Mermaid and Graphviz. This ensures that the criteria for node content, link direction, and rendering accuracy are consistent across both manual and automated methods.

\end{document}